# Network Motifs Analysis of Croatian Literature

Hana Rizvić, Sanda Martinčić-Ipšić, Ana Meštrović
Department of Informatics
University of Rijeka
Radmile Matejčić 2, Rijeka
{hrizvic, smarti, amestrovic}@uniri.hr

**Abstract:** *In this paper we analyse network motifs in the co-occurrence directed networks constructed from five different texts (four books and one portal) in the Croatian language. After preparing the data and network construction, we perform the network motif analysis. We analyse the motif frequencies and Z-scores in the five networks. We present the triad significance profile for five datasets. Furthermore, we compare our results with the existing results for the linguistic networks. Firstly, we show that the triad significance profile for the Croatian language is very similar with the other languages and all the networks belong to the same family of networks. However, there are certain differences between the Croatian language and other analysed languages. We conclude that this is due to the free word-order of the Croatian language.*

## 1 Introduction

Many scientists from different disciplines study networks because of their ubiquity. The complex networks in nature share global properties such as small-world property of short paths between vertices and highly clustered connections [16]. In addition, many of these networks are scale-free networks, characterised by power-law degree distribution [2]. However, besides these global network characteristics, there are certain properties on the meso-scale and local-scale [4] that explain structural differences between complex networks. That is why more detailed network analysis on the meso-scale and on the local-level is important. Network analysis on the meso-scale and local-scale may include: community detection [12], motif analysis [10] or graphlet analysis [13].

In this paper we are focused on the network motifs' analysis. Network motifs are connected and directed subgraphs occurring in complex networks at numbers that are significantly higher than those in randomized networks [10]. Motifs may contain up to 8 vertices. For now, there have been reports on 3-vertex and 4-vertex motifs due to the complexity of the algorithm that identifies the motifs from the complex networks.

Alon et al. [11] analyse superfamilies of networks based on significant motifs (Fig. 1.). The first group of networks are from three microorganisms: the Escherichia coli, Bacillus subtilis and the Saccharomyces cerevisiae. These microorganisms form sensory transcription networks, the vertices represent genes or operons and the edges represent direct transcriptional regulation. They form the first superfamily which includes three types of biological networks: signal-transduction interactions in mammalian cells, developmental transcription networks arising from the review of the development of the fruit fly and sea-urchin, and synaptic wiring between neurons in Caenorhabditis elegans. They also studied three WWW networks of hyperlinks between web pages related to university, literature and music. A feature of these networks is the transitivity of the relations, as evidenced by the motifs presented in these networks that are highly transitive. Similar results are obtained by testing three social networks, where people from the group are represented by vertices. The connections between two people, a positive opinion of one member of the group to another member, were represented by edges, obtained on the basis of questionnaires. The conclusion is that social networks and the web are probably members of the same superfamily, which may facilitate the understanding of the structure of the web. Furthermore, word-adjacency networks are analysed so that each vertex represented a single word, and each edge represented a connection between the two words that have followed one another in the text. The results obtained for different texts in different languages (English, French, Spanish and Japanese) are similar. Significant triads are from $ID_3$#1 to $ID_3$#6 (considering the IDs in [11]), and underrepresented are all other triads, from the $ID_3$#7 to $ID_3$#13. This means that the examined languages do not have a transitive relation such as the WWW. The explanation for these results may be in the structure of language, where words are divided into categories and generally the rule is that a word from one category follows a word from the other category. As an example, most connected category words are prepositions and behind them usually follows a noun or an article.

Biemann at al. [3] use motifs to quantify the differences between a natural and a generated language. The frequencies of three-vertex and four-vertex motifs for six languages are compared with artificially generated language from *n*-grams. An *n*-gram is contiguous sequence of *n* units (words) reflecting the statistical properties of a given text (or speech). The authors show that the four-vertex motifs can be interpreted by semantic relations of polysemy and synonymy of words.

This work has been supported in part by the University of Rijeka under the project number 13.13.2.2.07.

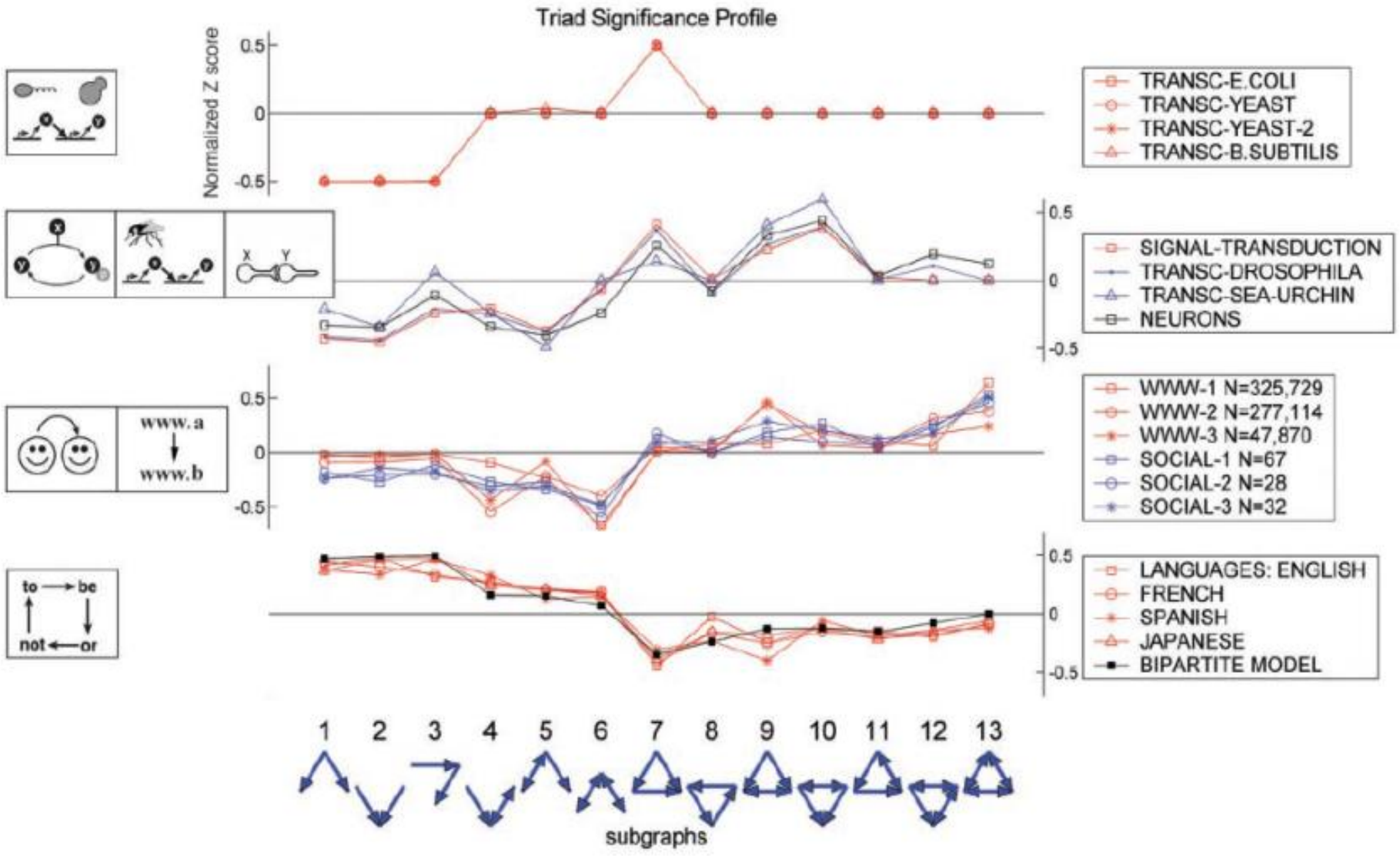

Figure 1: Superfamilies of complex networks according to the triad significance profile [11]

Our motivation for this research was to determine whether the local structure of the Croatian language networks share the same properties as other language networks. Croatian is a highly flective Slavic language and words can have seven different cases for singular and seven for plural, genders and numbers. The Croatian word order is mostly free, especially in non-formal writing. These features position Croatian among morphologically rich and free word-order languages. So far Croatian has been quantified in a complex networks framework based on the word co-occurrences [7], [1] and compared with shuffled counterparts [8], [9].

In this paper we describe the network motifs analysis of the co-occurrence directed networks constructed from the Croatian texts: four books and one forum. We use an approach based on the significance profile (*SP*) presented in [10]. We analyse three-vertex subgraphs called triads and present the results of triad significance profile (*TSP*) for the five analysed networks. In this paper we compare our results with *TSP* for other languages.

In the second section we give an overview of network motifs. In the third section we describe the experiment, and the fourth section presents the results. We conclude with some finishing remarks and the plans for future work.

## 2 Network motifs

A network motif is a small subgraph that appears more frequently in the real network than in the random network. The motif may be referred to as a significantly overrepresented subgraph in the network. As well, an underrepresented subgraph in the network is called an anti-motif. In [10] authors define network motifs as small patterns for which the probability of occurrence in a randomized network is less than the probability of occurrence in the real network with the cut-off value equal to 0.01.

In Fig. 2. are all 13 possible three-vertex connected directed subgraphs (triads). The triad ID notation in this paper is preserving the same notation as on the Fig 2 and it is the notation according to [11].

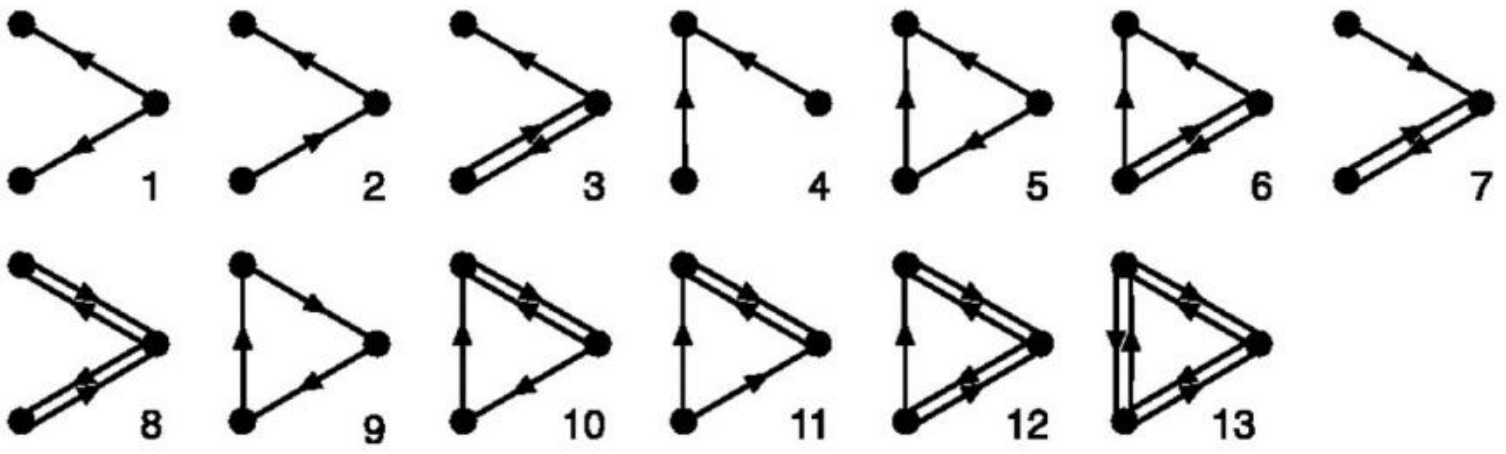

Figure 2: All 13 types of three-vertex connected subgraphs

In Fig. 3. are all 199 possible four-vertex connected directed subgraphs.

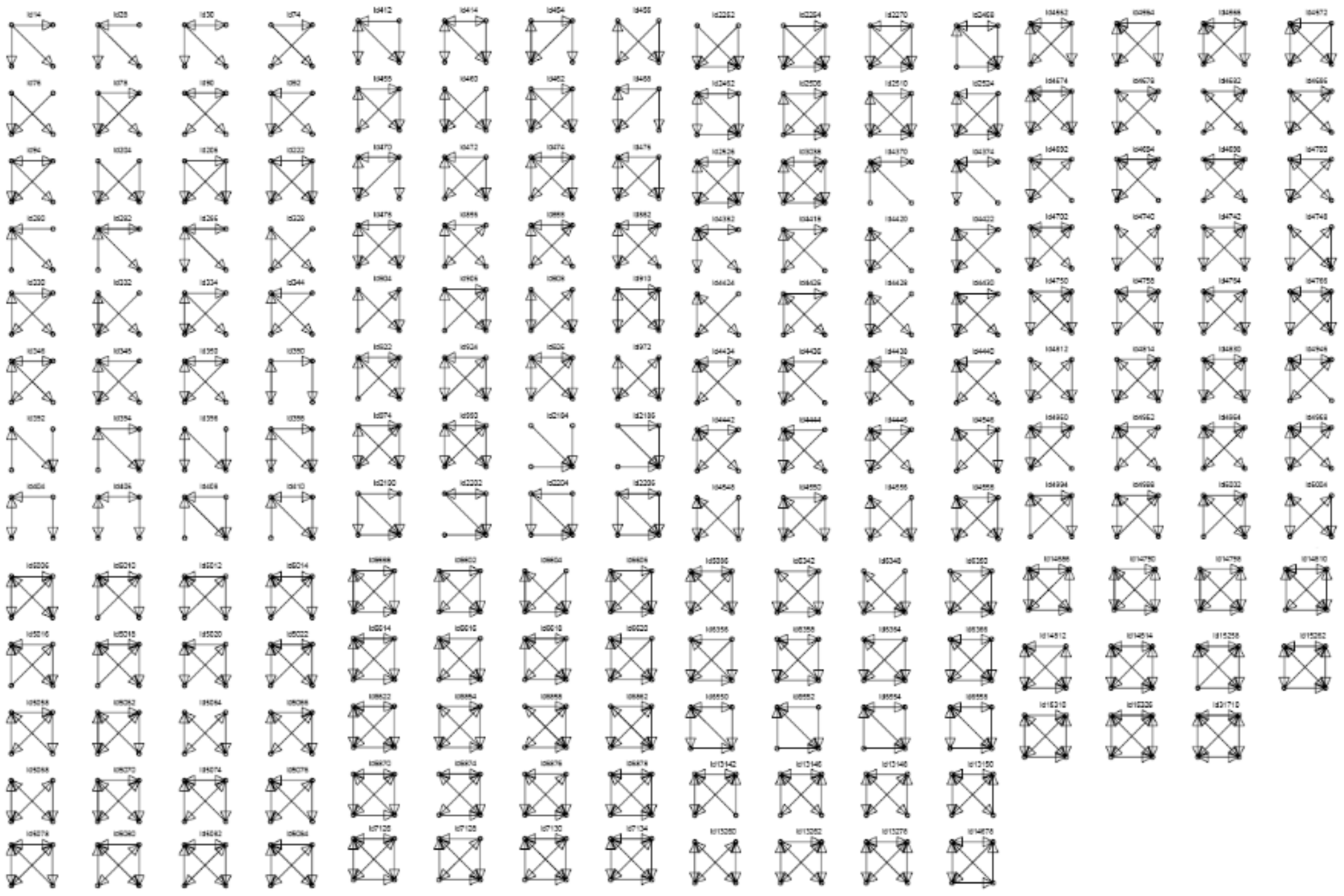

Figure 3: Four-vertex connected and directed subgraphs

Now, we will give the mathematical description of the motif in the graph or network *G*. There are two graphs (networks) *H* and *G* with non-empty sets of: vertices, edges and incidence relation. Let *H* be the real subgraph of *G*, $H \subset G$. The number of occurrences of graph *H* in graph *G*, we define as the frequency of *H* in *G*, written like $F_H(G)$. Some graph is frequent in *G* if its frequency in *G* is higher than cut-off value. Let $\Omega(G)$ be a family of randomized graphs of *G* (randomized graph has the same number of vertices and same degree sequence [10]). Now we choose *n* random graphs from $\Omega(G)$ uniformly, $R(G)$. Then we find out the frequency of the certain frequent sub-graph *H* in *G*. If the frequency of *H* in *G* is higher than its arithmetic mean frequency in *n* random graphs $R_i$, where $1 \leq i \leq n$, we call this sample significant and *H* is network motif for *G*.

Besides the frequency, motifs can be detected by using probabilities. The *p*-value of the motif is the number of random networks in which a particular motif appeared more frequently than in the original network, divided by the total number of generated random networks. Obviously, the *p*-value is between 0 and 1. The smaller the *p*-value of the motif is, the more significant the motif is.

Another measure for motif detection is a *Z*-score. The *Z*-score for the subgraph *H* in *G* can be calculated using the equation:

$$Z(H) = \frac{F_G(H) - \mu_R(H)}{\sigma_R(H)} \qquad (1)$$

where $\mu_R(H)$ is the mean and $\sigma_R(H)$ is the standard deviation of frequencies of *H* in the set of random graphs of *G*, $R(G)$. The higher the *Z*-score is, the more significant a detected motif is. Using eq. 1, for each subgraph *i*, we can calculate the statistical significance which is described as *Z*-score, $Z_i$.

Furthermore, the *SP* is the vector of *Z*-scores normalised to length 1:

$$SP_i = \frac{Z_i}{\sqrt{\sum Z_i^2}} \qquad (2)$$

# 3 Experiment

## 3.1 Datasets and networks construction

In our study, we examined five literary works. Our dataset contains five different texts; four books: *Mama Leone* (*ML*), *The Return of Philip Latinowicz* (*PL*), *The Picture of Dorian Gray*, (*DG*), *Bones*, (*BO*) and one forum *Narodne novine* (*NN*). All the books were written in or have been translated into Croatian. The web forum is selected as a representative of a different text genre in order to verify whether the observed properties are also valid for more relaxed genres besides those strictly for the literature.

The datasets are different in the size as well as in the size of the vocabulary (Table 1).

The texts were cleared of the index of contents, the authors' biographies and page numbers. Then we constructed directed co-occurrence networks (word-adjacency networks) in a way that each word represents a vertex, and the two words that follow one another establish the edge.

Table 1: Number of words, vertices and edges in the analysed datasets

| Dataset | Number of words | Number of vertices (*N*) | Number of edges (*K*) |
|---|---|---|---|
| *ML* | 86,043 | 12,416 | 52,012 |
| *PL* | 28,301 | 9,166 | 22,344 |
| *DG* | 75,142 | 14,120 | 47,823 |
| *BO* | 199,188 | 25,020 | 106,999 |
| *NN* | 146,731 | 13,036 | 55,661 |

### 3.2 Network motifs analysis

To analyse the motifs in networks we used the FANMOD tool [16]. FANMOD can search for motifs of three to eight vertices sizes using the rand-esu algorithm [15], which is much faster than similar tools, and the advantage is that it has a simple graphical interface and it is very intuitive to use.

The first step is the preparation of the input data: conversion of words to integers, where every number represents one vertex uniquely in the network, hence two integers in a line form an edge. Every line must contain at least two integers and a maximum of up to five integers. FANMOD provides the possibility to choose whether the networks have directed, undirected or coloured edges or vertices. We used directed uncoloured networks.

The algorithm options frame must be adjusted prior to running the algorithm itself. The options' frame includes: the set of the subgraph size and the setting of the switch between full enumeration and enumeration on a few samples. Motifs are identified through the comparison of frequencies in the original network and those in a random network so it is important to determine the number of random networks. It can be set up in the random networks frame in the box named 'Number of networks'. The default value for this is 1,000 networks but it can be increased if necessary. In this frame there are some important parameters: the parameter "exchanges per edge" (showing how many times the program goes through the edges) should be increased only if our results (output after the first reading) for a random network are very similar to the results for the original network. The parameter "exchange attempts" - if in the results there appears a small number of successful replacements, then we need to increase it, but it is important to bear in mind that if we have a few successful replacements it may mean that something is wrong with the network.

FANMOD produces results in terms of: *Z*-scores, *p*-values and frequencies. When we analyse the results, it is desirable to obtain as much as possible undefined *Z*-scores. If we have a lot of undefined *Z*-scores, it is not possible to determine which motif is significant (because the greater the Z-score is, the greater significance of this motif is). So if we have a lot of undefined *Z*-scores we have to increase the number of random networks, which will slow down the algorithm.

In the output file format is advisable to include an ASCII – text option for the easier reading of the results, and in HTML format for the presentation of the results. We calculate *Z*-scores for all triads in all five networks using FANMOD. After that we calculate *TPS* according to the eq. 2.

# 4 Results

The frequencies of all possible triads for five networks are presented in Fig. 4. In general, the triad frequencies behave similarly for all five networks. Therefore the Croatian language is comparable with other languages [3], [11]. Still, it is possible to identify differences between data source on $ID_3$#1 and $ID_3$#5 on the linear scale and $ID_3$#9, $ID_3$#11 and $ID_3$#13 on the logarithmic scale.

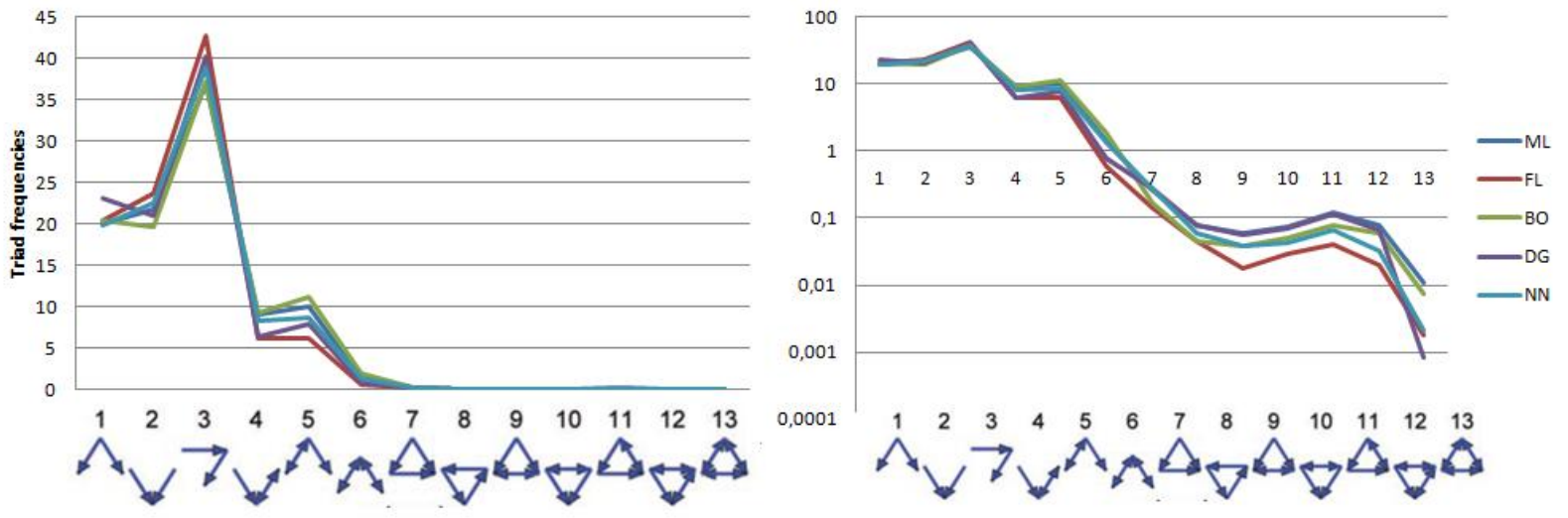

Figure 4: The frequencies of the triads for 5 datasets presented on the linear scale (left) and on the logarithmic scale (right)

Furthermore, we analyse *TSP* in order to detect which triads are significantly overrepresented (motifs) and which triads are significantly underrepresented (anti-motifs) and to compare it across the five different datasets. The results are presented in the diagram shown in the Fig. 5.

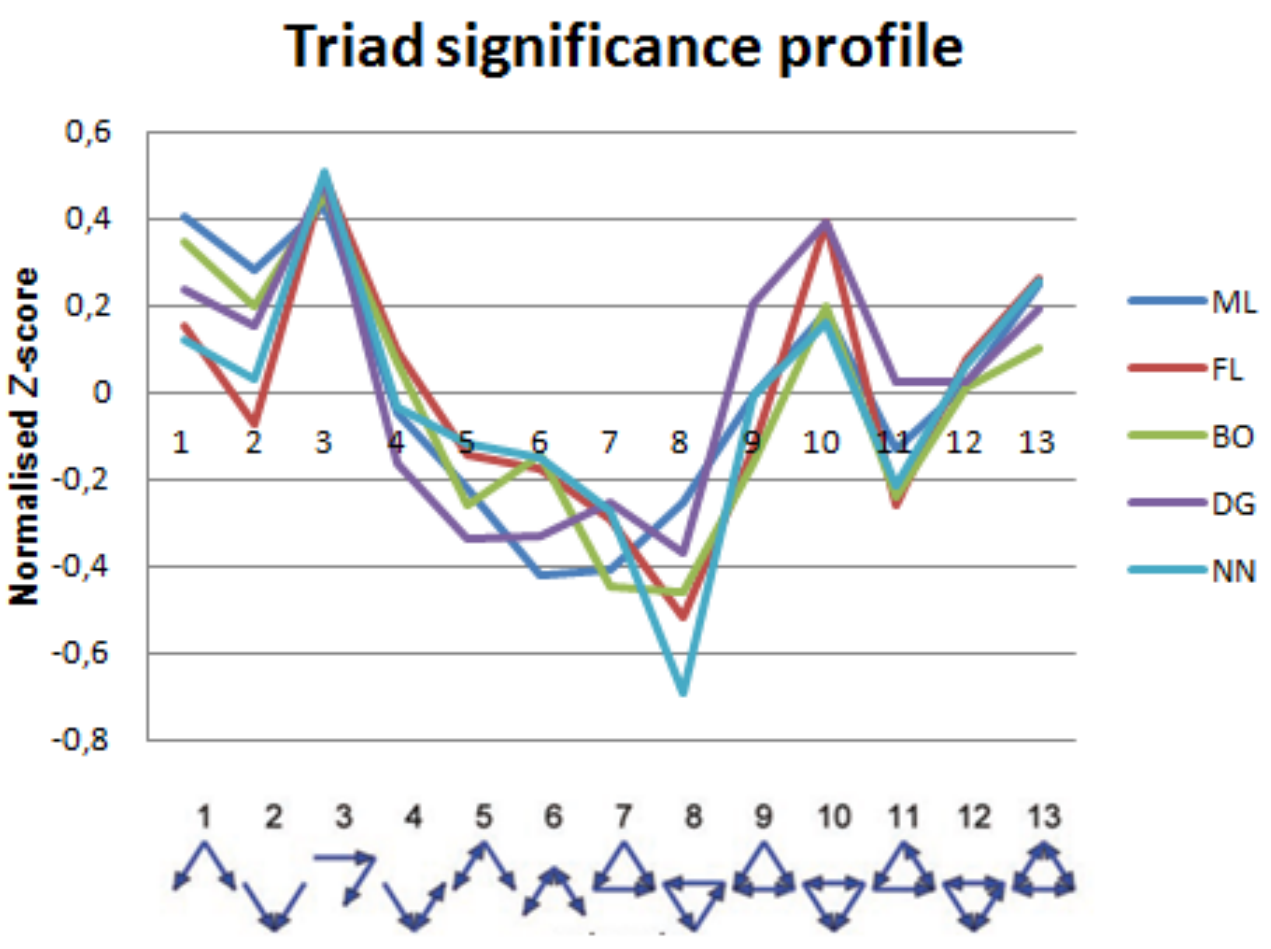

Figure 5: Triad significance profile for 5 datasets

There are several significantly overrepresented triads ($ID_3$#1, $ID_3$#3, $ID_3$#10 and $ID_3$#13). Triads with two edges ($ID_3$#1 and $ID_3$#3) are, based on the other reported results [3], [11], expected to be overrepresented in language networks. However, in our results, triads $ID_3$#10 and $ID_3$#13 are not likely to be overrepresented in language. It seems that this is inherent to languages with a free word-order such as Croatian. For example for three vertices of words: jako (*much*), ga (*him*), voli (*loves*); in Croatian language it is possible to have all six pairs of words (even triplets) as it is shown in Fig. 6. In opposite, in English language is impossible to have "him loves" as a part of the sentence.

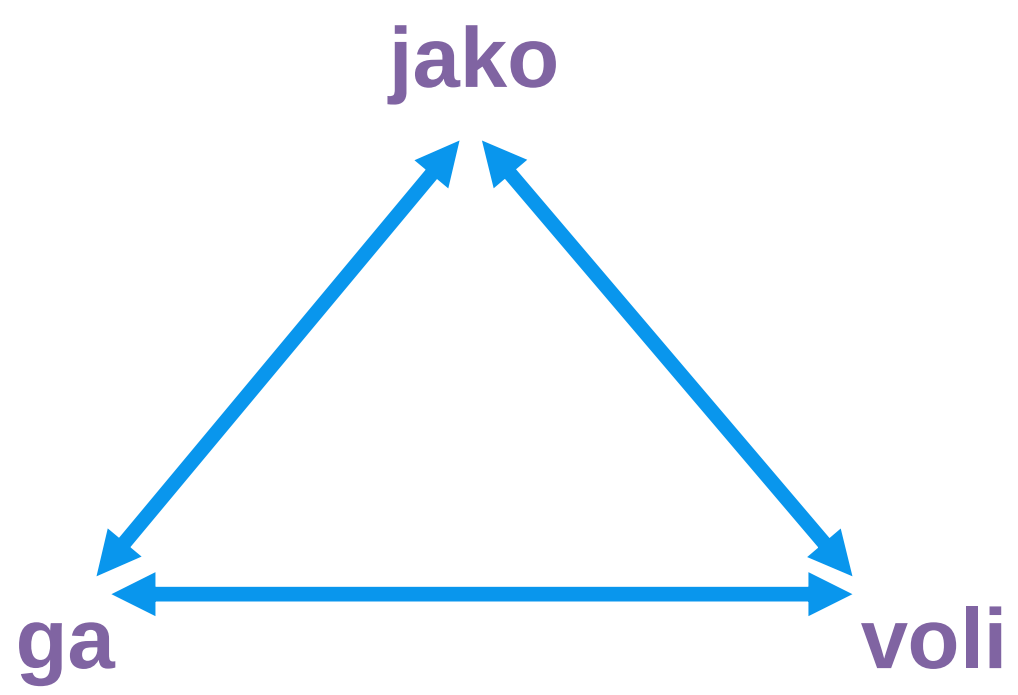

Figure 6: An example of the triad with $ID_3$#13 in Croatian language

# 5 Conclusion

In this paper we present the results of the network motifs analysis of Croatian literature. Motifs are used to detect structural similarities between directed networks of four books and one forum. We analyse triad significance profile in five different texts represented as directed co-occurrence networks.

The results show that Croatian language networks have similar triad significance profiles with other already analysed languages. Generally, in all language networks triads with two edges are overrepresented, while triads with three edges are underrepresented. For the Croatian language, there is an exception with three-edge triads ID3#10 and ID3#13 which are overrepresented. The overrepresentation of three-edge triads is caused by the free word-order nature of Croatian language.

It seems that motif-based analysis of the language networks is sensitive to the word order and syntax rules. And maybe it is possible to use it for the fine-grained differentiation of languages. Therefore, we will perform motif-based analysis of language networks for different languages. We will also include syntax networks and sub-word level networks (syllable networks, grapheme networks) in the analysis. Finally we plan to analyse the presence of the four-vertex motifs in language networks in order to see if they can be interpreted by the semantic relations in the polysemy and synonymy of words.

# 6 References

[1] Ban, K, Martinčić-Ipšić S., Meštrović, A. Initial comparison of linguistic networks measures for parallel texts. In 5th International Conference on Information Technologies and Information Society (ITIS), pp. 97-104. 2013.

[2] Barabási, A.- L., Albert, R. Emergence of scaling in random networks. Science 286, no. 5439 (1999): 509-512.
[3] Biemann, C., Roos S., Weihe, K. Quantifying semantics using complex network analysis. COLING12. 2012.
[4] Borge-Holthoefer, J., Arenas, A. Semantic networks: structure and dynamics. Entropy 12, no. 5 (2010): 1264-1302.
[5] Cancho, R. – F., Solé, R. The small world of human language. Proceedings of the Royal Society of London. Series B: Biological Sciences 268, no. 1482 (2001): 2261-2265.
[6] Hagberg, A., Swart, P., Chult, D. Exploring network structure, dynamics, and function using NetworkX. No. LA-UR-08-05495; LA-UR-08-5495. Los Alamos National Laboratory (LANL), 2008.
[7] Margan, D., Martincic-Ipšic, S., Meštrovic, A. Preliminary report on the structure of Croatian linguistic co-occurrence networks. 5th International Conference on Information Technologies and Information Society (ITIS), 89-96, 2013.
[8] Margan, D, Martincic-Ipšic, S., Meštrovic, A. Network Differences between Normal and Shuffled Texts: Case of Croatian. In Complex Networks V, pp. 275-283. Springer International Publishing, 2014.
[9] Margan, D., Meštrović, A., Martinčić-Ipšić, S. Complex Networks Measures for Differentiation between Normal and Shuffled Croatian Texts. In IEEE 37th International Convention on Information and Communication Technology, Electronics and Microelectronics (MIPRO 2014). 2014
[10] Milo, R.; Shen-Orr, S.; Itzkovitz, S. et al. Network motifs: simple building blocks of complex networks. Science, 298(5594): 824 – 827, 2002
[11] Milo, R; Itzkovitz, S; Alon, U. et al. Superfamilies of evolved and designed networks. Science, 303(5663): 1538 – 1542, 2004.
[12] Newman, Mark EJ. "The structure and function of complex networks." SIAM review 45, no. 2 (2003): 167-256.
[13] Pržulj, Nataša. "Biological network comparison using graphlet degree distribution." Bioinformatics 23, no. 2 (2007): e177-e183.
[14] Rasche, F; Wernicke, S. FANMOD fast network motif detection – manual, Bioinformatics, 22(9):1152–1153, 2006.
[15] Watts, Duncan J., and Steven H. Strogatz. Collective dynamics of ‘small-world’networks. nature 393, no. 6684 (1998): 440-442.
[16] Wernicke, S. A faster algorithm for detecting network motifs. In R. Cassadio and G. Myers, editors, Proceedings of WABI ’05, number 3692 in LNBI, pages 165–177. Springer-Verlag, 2005.